\documentclass{article}

\usepackage[preprint]{neurips_2026}

\usepackage[utf8]{inputenc}
\usepackage[T1]{fontenc}
\usepackage{url}
\usepackage{booktabs}
\usepackage{amsfonts}
\usepackage{amsmath}
\usepackage{nicefrac}
\usepackage{microtype}
\usepackage{xcolor}
\usepackage{graphicx}
\usepackage{float}
\usepackage{hyperref}
\definecolor{linkblue}{HTML}{1F5AA6}
\hypersetup{colorlinks=true, linkcolor=black, citecolor=black, urlcolor=linkblue}

\title{Warp-as-History: Generalizable Camera-Controlled Video Generation from One Training Video}

\author{%
Yifan Wang\textsuperscript{1,2} \quad Tong He\textsuperscript{2,3}\\
{\textsuperscript{1}Shanghai Jiao Tong University \quad
\textsuperscript{2}Shanghai AI Laboratory \quad
\textsuperscript{3}Shanghai Innovation Institute}\\[0.2em]
{\normalfont
Project Page: \href{https://yyfz.github.io/warp-as-history/}{https://yyfz.github.io/warp-as-history/}
\quad
Code: \href{https://github.com/yyfz/Warp-as-History}{https://github.com/yyfz/Warp-as-History}}
}

\begin{document}

\maketitle
\vspace{-2.8em}
\begin{figure}[H]
  \centering
  \includegraphics[width=\linewidth]{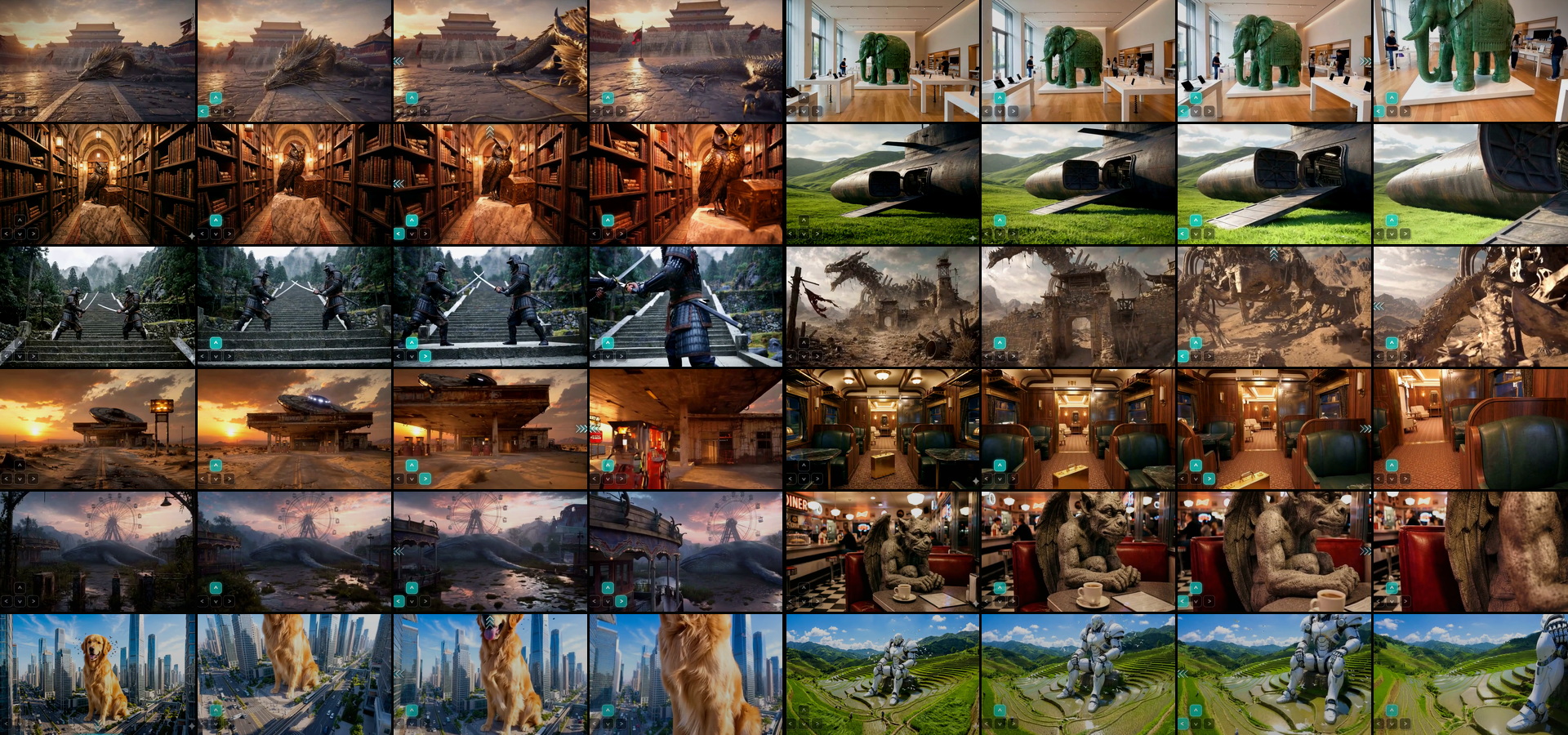}
  \vspace{-1.0em}
  \caption{\emph{Warp-as-History} generalizes to unseen scenes and unseen trajectories after finetuning on one video and one camera trajectory.}
  \label{fig:teaser}
  \vspace{-0.9em}
\end{figure}
\vspace{-1.0em}
\begin{abstract}
  Camera-controlled video generation has made substantial progress, enabling generated videos to follow prescribed viewpoint trajectories.
  However, existing methods usually learn camera-specific conditioning through camera encoders, control branches, or attention and positional-encoding modifications, which often require post-training on large-scale camera-annotated videos.
  Training-free alternatives avoid such post-training, but often shift the cost to test-time optimization or extra denoising-time guidance.
  We propose \emph{Warp-as-History}, a simple interface that turns camera-induced warps into camera-warped pseudo-history with target-frame positional alignment and visible-token selection.
  Given a target camera trajectory, we construct camera-warped pseudo-history from past observations and feed it through the model's visual-history pathway.
  Crucially, we align its positional encoding with the target frames being denoised and remove warped-history tokens without valid source observations.
  Without any training, architectural modification, or test-time optimization, this modification reveals a non-trivial zero-shot capability of a frozen video generation model to follow camera trajectories. Moreover, lightweight offline LoRA finetuning on only one camera-annotated video further improves this capability and generalizes to unseen videos, improving camera adherence, visual quality, and motion dynamics without test-time optimization or target-video adaptation.
  Extensive experiments on diverse datasets confirm the effectiveness of our method.
  \end{abstract}

\begin{figure}[t]
  \centering
  \includegraphics[width=\linewidth]{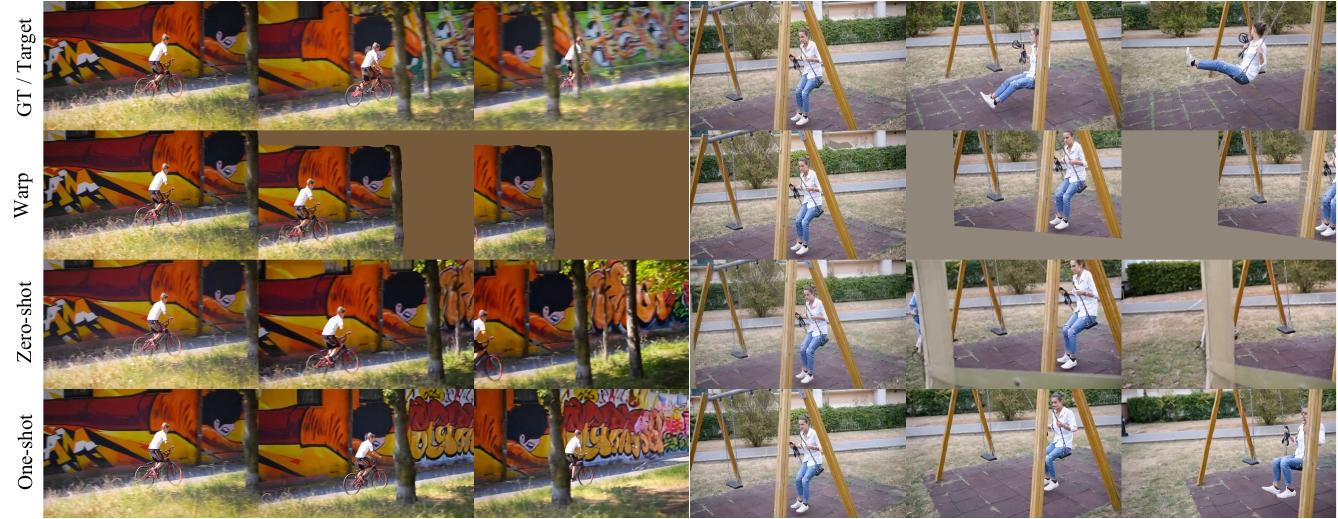}
  \vspace{-0.8em}
  \caption{From zero-shot history conditioning to one-training-video finetuning. Given the first image and a predefined camera trajectory, the four rows show ground truth, the camera-induced warp, zero-shot Warp-as-History, and one-training-video finetuning. The frozen model already turns the warp into visible camera-follow behavior through the pretrained history interface; one-training-video finetuning further stabilizes this behavior using a single separate training video, without test-time fitting to the shown video.}
  \label{fig:intro-zero-shot-observation}
  \vspace{-0.8em}
\end{figure}

\section{Introduction}

Camera motion is a primary control signal for interactive video generation.
It determines not only the viewpoint, but also which regions become visible, how objects move relative to the observer, and whether a generated scene can be explored beyond its initial frame.
This makes camera control in dynamic video more demanding than static novel-view synthesis: the model must enforce a prescribed camera trajectory while preserving appearance, disoccluding new content, and allowing foreground objects to move independently of the camera.

Recent progress in camera-controlled and interactive video generation has largely been driven by dedicated camera-control mechanisms.
Training-based methods inject camera information through camera encoders, control branches, attention or positional-encoding modifications, or related architectural changes, and typically require post-training on camera-annotated videos~\citep{he2024cameractrl,li2025cameras,zhang2025unified,ren2025gen3c,yu2024viewcrafter,huang2025voyager}.
Training-free methods avoid such post-training, but often enforce the desired trajectory at inference time through test-time optimization, denoising-time guidance, warp-and-repaint procedures, or other sampling-time constraints~\citep{hou2024training,you2024nvs,liu2024novel,zhou2025latent,song2025worldforge}.
At the same time, recent video generation models already exhibit surprisingly rich camera-motion behavior, suggesting that camera control may be latent in video generation models.
The challenge is therefore to expose and reliably steer this capability with minimal additional machinery, ideally without collecting large-scale camera-annotated videos, adding camera-specific modules, or imposing extra inference-time objectives.

We approach this question from the perspective of history-conditioned video generation.
Many video generation models already condition on visual history to continue a scene from previously observed frames.
This history pathway is usually treated as temporal context, but it is also a learned interface for interpreting appearance continuity, motion evidence, and incomplete observations.
We ask whether camera-induced geometric evidence can be presented through this existing interface.
Specifically, can warped observations induced by a target camera trajectory be used as history evidence, rather than as a dedicated adapter, camera-aware attention or positional encoding, or inference-time guidance objective?

Our answer is yes, when the geometric cue is expressed as history-conditioned evidence. We construct target-frame-aligned, visibility-aware warped observations: source-visible regions provide history evidence, while newly revealed regions are left to the pretrained generator for completion. Warping itself is not new; it appears in prior camera-control, view-synthesis, guidance, and repainting methods. Our distinction is where the warp enters generation: through the visual-history pathway, rather than as a sampling-time constraint or repainting signal.

Given the first frame and a pre-defined camera trajectory, Figure~\ref{fig:intro-zero-shot-observation} presents examples of: ground truth observation, the camera-induced warp, zero-shot Warp-as-History, and one-training-video finetuning.
The warp captures the prescribed camera-induced motion but remains an imperfect geometric cue.
When encoded through the pretrained history-conditioning pathway, this imperfect cue already elicits zero-shot camera-following capability from the frozen model, even in scenes with substantial foreground motion. Although this zero-shot effect is not robust enough to serve as a final method on its own, it reveals a useful latent capability: pretrained video generators can interpret camera-induced geometric evidence when provided as history-conditioned visual evidence.

This observation motivates \emph{Warp-as-History}, a low-resource camera-control framework rather than a new camera-conditioned video generation model. It keeps the control signal visual and geometry-aware, injecting it through the model's native history-conditioning pathway rather than converting it into a hard rendering target or an inference-time guidance objective. We further enhance this capability with lightweight offline LoRA finetuning on only one single camera-annotated video, improving camera adherence, visual quality, and motion dynamics without test-time optimization or target-video adaptation. Section~\ref{sec:method} gives the exact construction.

The same view also clarifies the role of finetuning.
If zero-shot Warp-as-History already produces measurable camera-follow behavior, lightweight finetuning can be studied as behavior stabilization: it adjusts when the model follows visible warp evidence, when it ignores unreliable warp regions, and when it relies on its generative prior for dynamics and disocclusion.
We use one-training-video finetuning as a diagnostic: when a single separate training video improves camera adherence on unrelated test videos, it supports the view that the proposed history interface exposes behavior partially supported by pretraining.
In our experiments, one-training-video finetuning makes the zero-shot behavior visibly clearer, as shown in Figure~\ref{fig:intro-zero-shot-observation}.
This update is trained offline on a video separate from the test videos; it is not test-time fitting, per-video optimization, or adaptation to the test instance.

Our contributions are:
\begin{itemize}
    \item We show that pretrained history-conditioned video models contain a weak camera-follow prior, and introduce Warp-as-History to expose it: target camera trajectories are converted into camera-warped pseudo-history with temporal alignment and visibility-aware evidence selection, allowing the frozen model to produce measurable zero-shot camera-following behavior through its native history pathway.
    \item We demonstrate one-training-video activation: offline LoRA finetuning on a single separate video stabilizes the exposed behavior and generalizes to unseen videos, supporting the view that finetuning amplifies an existing prior rather than learning camera control from scratch.
    \item Experiments on WorldScore, RE10K, and DAVIS show that Warp-as-History, after finetuning on only one separate video, is competitive with recent state-of-the-art camera-control baselines trained on orders of magnitude more data, with comparable camera adherence and strong visual-quality and consistency metrics.
\end{itemize}

\section{Related Work}

\paragraph{Camera-controlled video generation.}
Camera-controlled video generation has largely followed two routes.
Camera-matrix conditioning methods such as CameraCtrl~\citep{he2024cameractrl}, PRoPE~\citep{li2025cameras}, and UCPE~\citep{zhang2025unified} inject camera parameters through control branches, camera-aware attention, or positional encodings.
Warp- and geometry-conditioned methods such as Gen3C~\citep{ren2025gen3c}, ViewCrafter~\citep{yu2024viewcrafter}, and Voyager~\citep{huang2025voyager} instead provide target-view evidence through warps, geometry representations, or rendered views.
These methods provide strong trajectory control, but often rely on camera-aware modules, geometric representations, or large-scale camera-related training data.
Our goal is different: we ask whether an existing history-conditioned video generation model can read camera motion through its native video-history interface.

\paragraph{Training-free camera control.}
Training-free methods avoid camera-specific post-training and are therefore an important comparison class.
Examples include Training-free Camera Control~\citep{hou2024training}, NVS-Solver~\citep{you2024nvs}, video-diffusion-prior novel-view extrapolation~\citep{liu2024novel}, Latent-Reframe~\citep{zhou2025latent}, and WorldForge~\citep{song2025worldforge}.
Many such methods still pay for control at inference time through test-time optimization, denoising guidance, latent repainting, recursive rollout, or related sampling-time procedures.
Warp-as-History instead constructs camera-induced history once and then follows the native sampler, without per-sample optimization or extra denoising-time guidance.

\paragraph{History-conditioned video generation.}
History-conditioned video generation uses previous frames as visual context for predicting future frames.
Recent methods~\citep{song2025history,huang2025self,yu2025context,wu2025video} explore how visual history and retrieved context can improve generation, rollout behavior, and scene consistency.
Helios~\citep{yuan2026helios} is a recent state-of-the-art history-conditioned backbone with a native history interface.
We build on this interface but change its role: history is no longer only temporal context, but an aligned camera-control signal.

\section{Method}
\label{sec:method}
\subsection{Overview: one-training-video Warp-as-History}

We first describe the pretrained interface that our method will reuse.
Write a video as $X=(x_1,\ldots,x_T)$ and let $p$ be the text prompt.
Let $p_\theta(\cdot\mid\cdot)$ denote the conditional sampling distribution induced by the pretrained history-conditioned video generation model and its sampling procedure.
For a chunk starting at time $t$, $X_{<t}$ denotes the available past frames and $X_{t:t+K}$ denotes the future chunk generated by the backbone.
The model consumes history through its native construction operator $\mathcal{H}$, which selects, encodes, and temporally packs past visual evidence into a history condition $H_t$.
In history-conditioned video generation, this history may be processed by a transform $\eta_t$ that corrupts, masks, or drops parts of the past.
With this notation, the model predicts future chunks from visual history:
\begin{equation}
    \label{eq:history-conditioned-history}
    \begin{aligned}
        \bar{X}_{<t} &= \eta_t(X_{<t}),\\
        H_t &= \mathcal{H}(\bar{X}_{<t}),\\
        X_{t:t+K} &\sim p_\theta(\cdot \mid H_t,p).
    \end{aligned}
\end{equation}
This notation highlights the interface we reuse: the model receives processed visual history through $H_t$ and samples the next chunk conditioned on that history and the text prompt.

\begin{figure}[t]
  \centering
  \includegraphics[width=\linewidth]{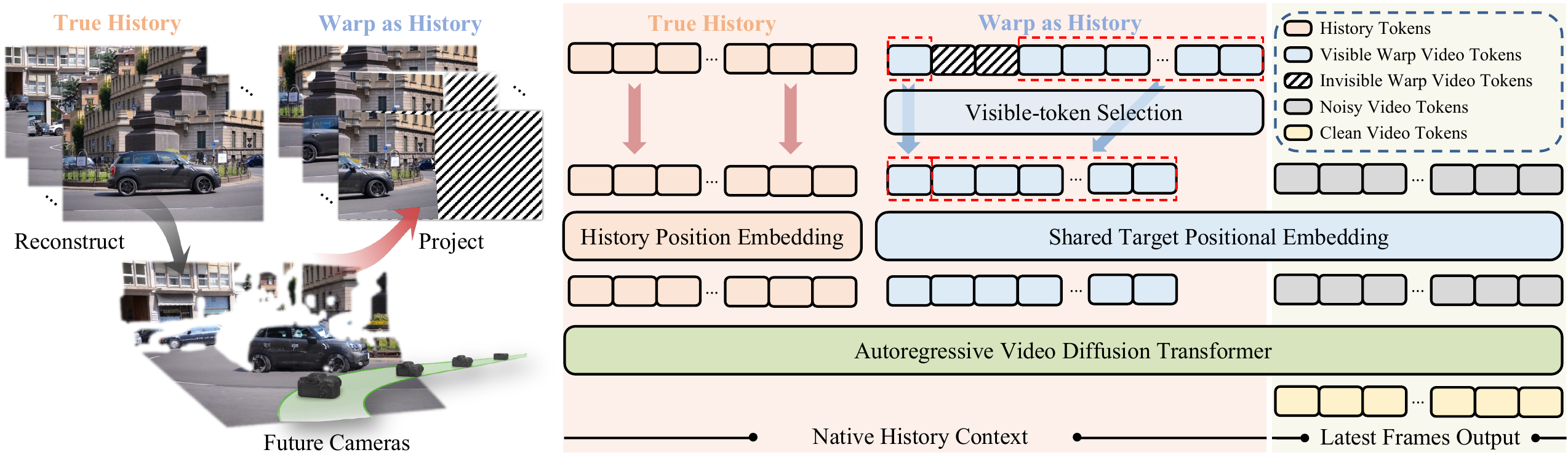}
  \vspace{-0.6em}
  \caption{Conditioning a video diffusion model on camera motion. Warp-as-History packs camera-warped pseudo-history into the native history stream, aligns it to target-frame positions, and applies visible-token selection.}
  \label{fig:method-overview}
  \vspace{-0.8em}
\end{figure}

Warp-as-History is the conditioning interface used by our one-training-video method.
It converts a target camera trajectory into camera-warped pseudo-history and feeds it through the native history pathway, with target-frame positional alignment and visible-token selection.
Applied directly to the frozen model, the same interface produces the zero-shot behavior discussed in the introduction; we use this behavior as diagnostic evidence that pretrained history-conditioned models can read camera-induced visual evidence from history.
The final model uses offline LoRA finetuning on one separate camera-annotated video to stabilize this behavior and improve quality, foreground dynamics, and disocclusion.
The resulting weights are shared across test videos; no test-time fitting or per-video optimization is used.
Figure~\ref{fig:method-overview} illustrates how Warp-as-History conditions the video diffusion model on camera motion.

\subsection{Warp-as-History conditioning}
\label{sec:warp-as-history}

We first define the conditioning interface that turns camera geometry into visual history evidence.
Geometric warps and rendered target views are already common camera-control signals; our use of a warp is not the novelty.
The design question is how a history-conditioned video generation model should receive such a signal without a new control branch, a learned camera encoder, or a sampling-time optimization loop.
We answer by converting the warp into the same kind of visual evidence the pretrained model already consumes as history, then aligning and filtering that evidence, as summarized in Figure~\ref{fig:method-overview}.

\paragraph{Camera-warped pseudo-history.}
Let $C=(c_1,\ldots,c_T)$ be the target camera trajectory for the generated window.
A camera-induced warp video $W_C$ renders the available observation under the target camera trajectory, producing an image-space camera-motion cue.
We first reconstruct the scene with an off-the-shelf reconstruction model~\citep{wang2025pi}, then project the reconstruction to each target camera in $C$ to obtain a 2D warp video.
Using it as a hard render target would encourage copying warp errors, while learning a new warp-conditioning branch would require extra camera-specific training.
We therefore route the warp through the native history interface, corresponding to the warp construction and history-packing path in Figure~\ref{fig:method-overview}.
Let $\tilde{H}^C_t$ denote the camera-warped pseudo-history condition:
\begin{equation}
	    \label{eq:pseudo-history}
	    \tilde{H}^{C}_t
	    =
	    \mathcal{S}_{M_C}(\mathcal{H}(W_C)).
\end{equation}
Here $M_C$ is the warp validity mask and $\mathcal{S}_{M_C}$ is visible-token selection applied after native history construction; $\mathcal{H}$ itself does not take a mask input.
The construction $\mathcal{H}$ is the same native history construction used by the backbone: the warped frames are patchified, encoded, and packed as ordinary visual history.
This condition differs from ordinary history only in how these history tokens are temporally positioned and in which tokens are retained as valid evidence.
With ordinary history placement, the warp is presented as past visual context, so the frozen model can apply its pretrained history-to-future continuation behavior to the camera-induced motion.
On the frozen model, this also serves as a diagnostic interface: if the base model can continue camera-induced visual motion from history, this condition should produce measurable camera-follow behavior before finetuning.
Section~\ref{sec:zero-shot-analysis} tests this zero-shot behavior directly.

\paragraph{Target-frame positional alignment.}
The pretrained continuation behavior separates past history from the current noisy chunk through both the history patchification path and temporal rotary positional embedding (RoPE) positions~\citep{su2024roformer}.
If warp tokens keep ordinary history positions, they remain valid history evidence for a motion trace to continue, but the $j$-th warp frame is still interpreted as past context rather than evidence for the $j$-th frame being denoised.
We therefore keep the warp in the history patchification path, but give each warp latent the same temporal position as the corresponding current noisy latent by assigning it the RoPE index of the target latent at the same frame order, as shown by the shared target positional embedding in Figure~\ref{fig:method-overview}.
Because these tokens are still inserted as history evidence, this remapping does not replace or overwrite the noisy target tokens.
Empirically, it is critical: normal denoising remains stable, and Figure~\ref{fig:interface-analysis} shows that the zero-shot output immediately starts to follow the warp after target-frame alignment.
The same effect also makes unreliable or invisible warp regions easier to copy, motivating the visible-token selection described next.

\paragraph{Visible-token selection.}
Camera motion creates newly visible areas that a first-frame warp cannot observe, and imperfect geometry can produce holes, stretched textures, or unreliable regions.
Rather than adding a separate conditioning input for the invisible mask, we make invalid evidence resemble the incomplete histories seen during history-conditioned pretraining.
Dropping invisible warp tokens from the DiT history stream, shown as visible-token selection in Figure~\ref{fig:method-overview}, leaves disocclusions to the pretrained completion behavior while still using reliable warped regions as camera-motion evidence.
In practice, the warp validity mask is mapped to the latent-token grid, and tokens with insufficient valid support are removed from the history stream.
Figure~\ref{fig:interface-analysis} shows the resulting zero-shot jump: after visible-token selection, the frozen model follows the target camera while completing regions that were invisible in the warp.
Section~\ref{sec:zero-shot-analysis} ablates this chain, and Figure~\ref{fig:intro-zero-shot-observation} shows the same behavior in the main qualitative example.
The behavior is still imperfect: the model may over-copy warped dynamic objects and produce unnatural boundaries near visibility changes, motivating the one-training-video finetuning used by the final model.

This pseudo-history can coexist with ordinary history:
\begin{equation}
    \label{eq:wah-conditioning}
    \hat{X}_{t:t+K}
    \sim
    p_\theta(\cdot \mid H_t,\tilde{H}^C_t,p).
\end{equation}
Both $H_t$ and $\tilde{H}^C_t$ are inserted through the model's native history stream; no new camera branch or sampling-time guidance loss is introduced.
In the first-window setting or ablations without clean history, $H_t$ can be empty.
We use this expression only to state the conditioning interface: camera control is represented by populating the existing history pathway with camera-warped pseudo-history.
The same conditioning form is used by the frozen-model diagnostic and by the one-video finetuned model.

\subsection{One-training-video LoRA finetuning}

The final Warp-as-History model keeps the conditioning interface above and finetunes the backbone with a lightweight LoRA update on one separate camera-annotated video.
The frozen-model diagnostic reveals camera-follow behavior, but does not solve all aspects of dynamic video generation.
In practice, the frozen model can still over-trust the camera-warped pseudo-history: dynamic foreground objects can be copied too rigidly from the warp, and visibility boundaries can remain unnatural.
We therefore use lightweight LoRA~\citep{hu2022lora} finetuning as the adaptation step.

The goal of finetuning is not to learn a new camera-control branch.
Instead, it adjusts how the pretrained history reader balances two sources of evidence: the visible warp tokens, which provide camera-induced motion cues, and the model's generative prior, which is needed for independent dynamics and disocclusion completion.
The training loss is the same video-generation objective used by the backbone; only the low-rank update is optimized.
The role of the update is to mitigate zero-shot artifacts and reduce the remaining distribution shift from natural histories $H_t$ to camera-warped pseudo-history $\tilde{H}^C_t$.

One-training-video finetuning is treated as a diagnostic for low-resource finetuning.
If a single held-out training video improves camera adherence across unrelated test videos, it suggests that the history-conditioning interface is exposing behavior already partially supported by the pretrained model.
Which single videos are effective for this finetuning is an empirical question, not part of the method definition; Section~\ref{sec:oneshot-source-analysis} studies it explicitly.
We treat additional training videos as a sensitivity check rather than a main method claim; Section~\ref{sec:few-shot-sensitivity} and Appendix~\ref{app:additional-data-sensitivity} report the current multi-video setting.

\subsection{Implementation details}
All experiments in this paper are built on Helios~\citep{yuan2026helios}, a real-time long-video generation model with native history conditioning.
Unless otherwise stated, the zero-shot experiments use the distilled Helios checkpoint.
The main recipe keeps the adaptation localized: aligned warp history and LoRA are inserted only in the first, lowest-resolution Helios stage, while later stages use the native refinement path.
The training loss is unchanged from the backbone, and inference uses the standard sampler without test-time optimization or extra denoising-time guidance.
In our runs, one-training-video LoRA finetuning uses 1000 iterations and takes about one hour on a single A800 GPU, already producing useful camera-control behavior when mounted on the distilled inference model.
Once the warp video is available, inserting Warp-as-History adds less than one second of overhead for generating a 33-frame chunk, since it only packs the camera-warped history condition and does not introduce an optimization loop or extra denoising-time guidance.
Appendix~\ref{app:full-interface-ablation-tables} provides the checkpoint, LoRA, and packing details used in the experiments.

\begin{figure}[t]
  \centering
  \includegraphics[width=\linewidth]{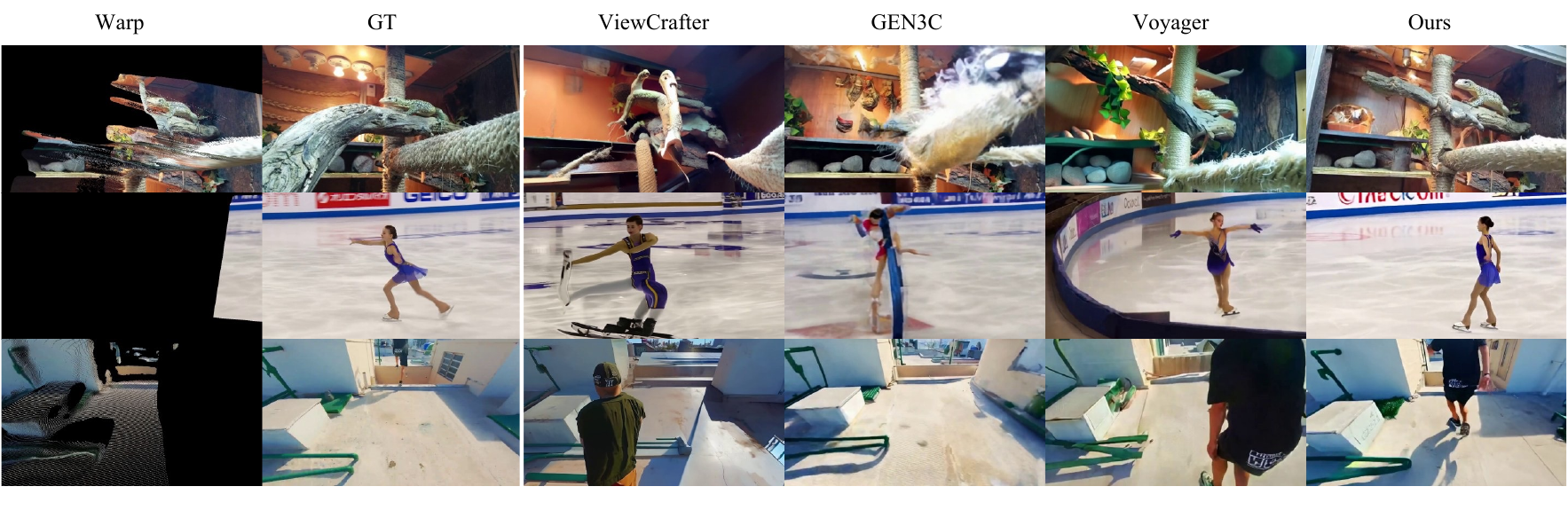}
  \vspace{-0.6em}
  \caption{Qualitative comparison with external camera-control methods on in-the-wild videos. Columns show the camera-induced warp, ground truth, ViewCrafter, Gen3C, Voyager, and ours under the same target camera setting.}
  \label{fig:external-qualitative-comparison}
  \vspace{-0.8em}
\end{figure}

\section{Analysis and Experiments}

The experiments are grouped by the claim they test.
First, we compare against prior camera-control systems on the public benchmarks used for static and dynamic video evaluation.
Second, we ask whether the frozen model can be induced to follow camera motion at all, and which interface choices make this behavior appear.
Third, we analyze how the choice and amount of small finetuning data affect activation of the camera-follow prior.

\subsection{Evaluation Datasets}

We evaluate on WorldScore~\citep{duan2025worldscore}, RealEstate10K (RE10K)~\citep{zhou2018stereo}, and DAVIS~\citep{perazzi2016benchmark}.
WorldScore provides a static world-generation benchmark, RE10K provides real static scenes with camera motion, and DAVIS provides dynamic videos with foreground motion.
Unless otherwise specified, \textit{Ours (one-shot)} denotes a single LoRA finetuning run on the DAVIS \emph{car-roundabout} video, evaluated without per-test-video adaptation.
The main text reports compact metrics for camera adherence, quality, consistency, and dynamics, including DOVER~\citep{wu2023exploring} and VBench~\citep{huang2024vbench} axes; full metric columns and evaluation details are provided in Appendix~\ref{app:full-interface-ablation-tables}.

\subsection{Comparison on Diverse Benchmarks}

The external comparisons use our one-shot setting unless otherwise stated, while full report cards are deferred to the appendix.

\paragraph{WorldScore comparison.}
Table~\ref{tab:worldscore-video-models} reports representative WorldScore~\citep{duan2025worldscore} results together with our Helios evaluations.
Helios-Distilled and Ours rows report native 33-frame full-static WorldScore results.

\begin{table}[H]
\caption{WorldScore results. Helios-Distilled and Ours rows report native 33-frame full-static WorldScore evaluation. Best values are bolded and second-best values are underlined.}
  \label{tab:worldscore-video-models}
  \centering
  \scriptsize
  \setlength{\tabcolsep}{3pt}
  \resizebox{\linewidth}{!}{%
  \begin{tabular}{lcccccccc}
    \toprule
    Method & Avg. & Camera & Object & Content & 3D & Photo. & Style & Subjective \\
    & & Control & Control & Align. & Cons. & Cons. & Cons. & Quality \\
    \midrule
    CogVideoX-I2V~\citep{yang2024cogvideox} & 62.15 & 38.27 & 40.07 & 36.73 & 86.21 & 88.12 & 83.22 & \underline{62.44} \\
    Voyager~\citep{huang2025voyager} & \underline{77.62} & \textbf{85.95} & \underline{66.92} & \textbf{68.92} & 81.56 & 85.99 & 84.89 & \textbf{71.09} \\
    FantasyWorld-1.0~\citep{dai2025fantasyworld} & \textbf{80.45} & \underline{81.45} & \textbf{87.90} & \underline{66.94} & 84.62 & \textbf{94.07} & 86.69 & 61.46 \\
    \midrule
    Helios-Distilled (text-only) & 62.42 & 26.42 & 42.66 & 37.75 & \textbf{92.54} & \underline{93.93} & \underline{90.41} & 53.21 \\
    Ours (zero-shot) & 63.26 & 61.32 & 33.07 & 39.92 & 87.27 & 88.18 & 85.67 & 47.37 \\
    Ours (one-shot) & 65.64 & 62.00 & 32.82 & 38.60 & \underline{89.36} & 90.43 & \textbf{91.46} & 54.83 \\
    \bottomrule
  \end{tabular}
  }
\end{table}

On WorldScore, Warp-as-History substantially increases camera controllability over the text-only Helios-Distilled baseline.
Camera Control rises from 26.42 to 61.32 in the zero-shot setting and 62.00 after one-shot finetuning, corresponding to relative gains of 132.1\% and 134.7\%, respectively.
One-shot finetuning mainly improves visual quality over the zero-shot interface: Subjective Quality increases from 47.37 to 54.83, a 15.7\% relative gain, while the average score also improves from 63.26 to 65.64.

\begin{figure}[H]
  \centering
  \vspace{-0.8em}
  \includegraphics[width=\linewidth]{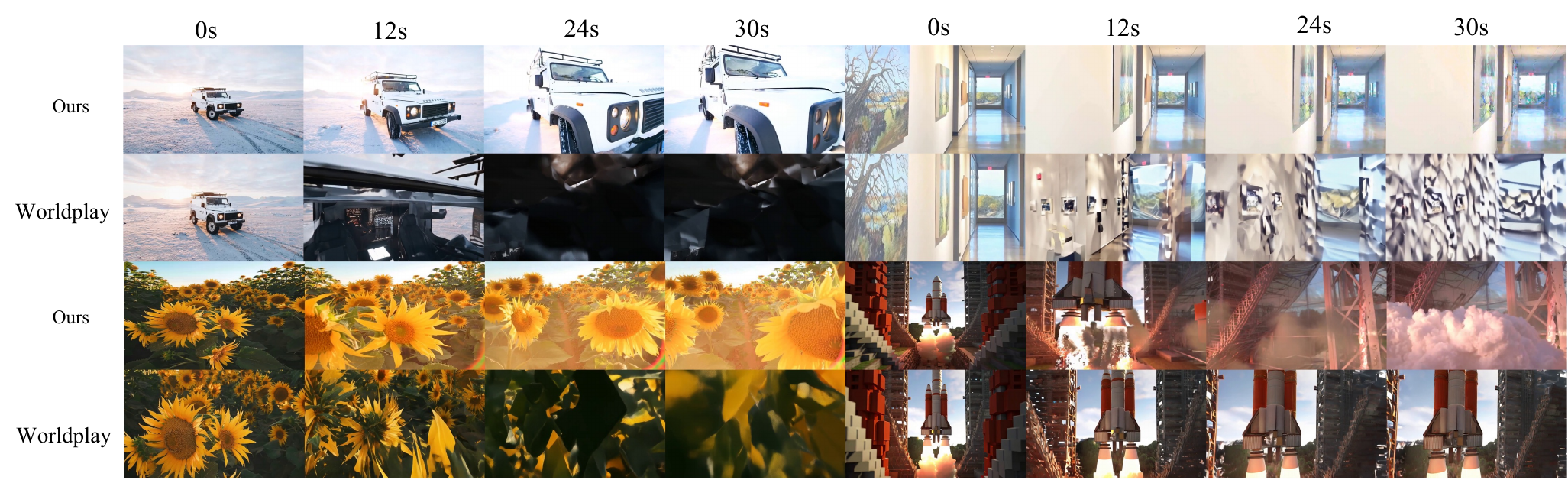}
  \vspace{-0.8em}
  \caption{Qualitative comparison with HyWorldPlay on 30-second trajectories sampled from WorldScore images. Frames are shown at 0, 12, 24, and 30 seconds.}
  \label{fig:hyworldplay-qualitative}
  \vspace{-0.8em}
\end{figure}

\paragraph{Long-video comparison with HyWorldPlay.}
Table~\ref{tab:hyworldplay-vbench} and Figure~\ref{fig:hyworldplay-qualitative} compare with HyWorldPlay~\citep{sun2025worldplay} on 30-second WorldScore-sampled trajectories; protocol details are in Appendix~\ref{app:hyworldplay-details}.
Ours is slightly better on Flicker and Motion Smoothness, while HyWorldPlay is stronger on VBench Overall, Imaging Quality, Dynamic Degree, and scene consistency.

\begin{table}[H]
  \caption{VBench comparison with HyWorldPlay on 30-second WorldScore-sampled trajectories.}
  \label{tab:hyworldplay-vbench}
  \centering
  \scriptsize
  \setlength{\tabcolsep}{3.2pt}
  \resizebox{\linewidth}{!}{%
  \begin{tabular}{@{}lccccccc@{}}
    \toprule
    Method & Flicker $\uparrow$ & Motion $\uparrow$ & Subject $\uparrow$ & Background $\uparrow$ & Dynamic $\uparrow$ & Imaging $\uparrow$ & Overall $\uparrow$ \\
    \midrule
    Ours & \textbf{0.984301} & \textbf{0.994419} & 0.850141 & 0.908713 & 0.340000 & 54.5424 & 0.770500 \\
    HyWorldPlay~\citep{sun2025worldplay} & 0.979760 & 0.990693 & \textbf{0.862521} & \textbf{0.918601} & \textbf{0.380000} & \textbf{62.8726} & \textbf{0.793383} \\
    \bottomrule
  \end{tabular}
  }
\end{table}

\paragraph{RE10K and DAVIS comparisons.}
Table~\ref{tab:external-geometry} reports camera-following metrics on RE10K and DAVIS, and Table~\ref{tab:external-quality} reports visual quality, temporal consistency, and dynamics.
Within each dataset, all methods use the same evaluation protocol; the exact subset construction is deferred to Appendix~\ref{app:full-interface-ablation-tables}.
The external baselines are Gen3C~\citep{ren2025gen3c}, Voyager~\citep{huang2025voyager}, and ViewCrafter~\citep{yu2024viewcrafter}.

\begin{table}[H]
  \caption{Camera-following evaluation on DAVIS and RE10K.}
  \label{tab:external-geometry}
  \centering
  \scriptsize
  \setlength{\tabcolsep}{2.8pt}
  \resizebox{\linewidth}{!}{%
  \begin{tabular}{@{}lllcccccc@{}}
    \toprule
    Dataset & Method & Training scale & PSNR $\uparrow$ & SSIM $\uparrow$ & LPIPS $\downarrow$ & Vis. LPIPS $\downarrow$ & R-Err $\downarrow$ & T-Err $\downarrow$ \\
    \midrule
    DAVIS & Gen3C & $\sim$90K videos & \textbf{16.29} & \textbf{0.5267} & \textbf{0.3539} & \textbf{0.1930} & \textbf{2.24} & \textbf{0.0663} \\
    & Voyager & $\sim$78K videos $\to$ $\sim$100K clips & 14.75 & 0.3983 & 0.4431 & 0.2558 & 3.05 & \underline{0.0706} \\
    & ViewCrafter & $\sim$85K videos $\to$ $\sim$632K clips & 14.72 & \underline{0.4133} & 0.3925 & 0.2308 & 3.85 & 0.1031 \\
    & Ours (one-shot) & 1 video $\to$ 4 clips & \underline{15.21} & 0.3976 & \underline{0.3794} & \underline{0.2236} & \underline{2.97} & 0.0942 \\
    \midrule
    RE10K & Gen3C & $\sim$90K videos & \textbf{20.10} & \textbf{0.7775} & \textbf{0.1523} & \textbf{0.0828} & \textbf{0.62} & \textbf{0.0158} \\
    & Voyager & $\sim$78K videos $\to$ $\sim$100K clips & \underline{19.03} & \underline{0.6914} & \underline{0.2304} & \underline{0.1268} & 0.86 & 0.0322 \\
    & ViewCrafter & $\sim$85K videos $\to$ $\sim$632K clips & 15.86 & 0.6765 & 0.2636 & 0.2015 & \underline{0.83} & \underline{0.0237} \\
    & Ours (one-shot) & 1 video $\to$ 4 clips & 17.15 & 0.6214 & 0.2343 & 0.1426 & 1.28 & 0.0454 \\
    \bottomrule
  \end{tabular}
  }
\end{table}

\begin{table}[H]
  \caption{Visual quality, temporal consistency, and dynamics on DAVIS and RE10K.}
  \label{tab:external-quality}
  \centering
  \scriptsize
  \setlength{\tabcolsep}{2.4pt}
  \resizebox{\linewidth}{!}{%
  \begin{tabular}{@{}llccccccccc@{}}
    \toprule
    Dataset & Method & FID $\downarrow$ & FVD $\downarrow$ & DOVER $\uparrow$ & Flicker $\uparrow$ & Motion $\uparrow$ & Subject $\uparrow$ & Backgr. $\uparrow$ & Dynamic $\uparrow$ & Imaging $\uparrow$ \\
    \midrule
    DAVIS & Gen3C & \underline{72.71} & \underline{64.98} & 0.326 & \underline{0.946} & \underline{0.977} & \underline{0.926} & \underline{0.936} & 0.701 & 58.06 \\
    & Voyager & 90.68 & 80.28 & 0.371 & \textbf{0.948} & 0.976 & 0.925 & 0.930 & \underline{0.727} & 58.61 \\
    & ViewCrafter & 76.54 & 65.76 & \textbf{0.507} & 0.933 & 0.961 & 0.885 & 0.903 & \textbf{0.818} & \textbf{67.33} \\
    & Ours (one-shot) & \textbf{68.18} & \textbf{57.95} & \underline{0.448} & 0.941 & \textbf{0.977} & \textbf{0.941} & \textbf{0.940} & 0.714 & \underline{62.20} \\
    \midrule
    RE10K & Gen3C & \textbf{16.55} & \textbf{12.66} & 0.298 & 0.951 & 0.989 & 0.917 & \underline{0.949} & \textbf{1.000} & 58.60 \\
    & Voyager & 30.65 & 22.49 & 0.382 & \textbf{0.963} & \textbf{0.992} & \underline{0.955} & 0.947 & 0.960 & 56.80 \\
    & ViewCrafter & 30.67 & 23.44 & \underline{0.436} & 0.947 & 0.983 & 0.881 & 0.935 & \textbf{1.000} & \underline{63.44} \\
    & Ours (one-shot) & \underline{22.11} & \underline{17.92} & \textbf{0.442} & \underline{0.955} & \underline{0.991} & \textbf{0.956} & \textbf{0.958} & \underline{0.980} & \textbf{65.97} \\
    \bottomrule
  \end{tabular}
  }
\end{table}

RE10K is a domain-stress test for our low-resource setting: our one-shot update is finetuned on DAVIS rather than on RE10K, whereas the external baselines use large-scale training data from the same real-estate domain.
Even under this mismatch, Ours remains in a comparable camera-following range and obtains stronger visual quality, including the best DOVER, Subject Consistency, Background Consistency, and Imaging scores.
On DAVIS, the visual-quality advantage is more pronounced: Ours has the best FID/FVD, Subject Consistency, and Background Consistency, while maintaining camera-following accuracy comparable to the external baselines.

\paragraph{Qualitative comparison.}
Figure~\ref{fig:external-qualitative-comparison} compares our one-training-video setting against the three external baselines on seven in-the-wild videos.
Together with the DAVIS quantitative results, these examples show that Warp-as-History preserves scene content and foreground motion more cleanly than prior warp-based baselines, which often expose warp artifacts, blur, or distorted objects.

\subsection{Ablating Warp-as-History}
\label{sec:zero-shot-analysis}
\label{sec:interface-analysis}

\begin{figure}[H]
  \centering
  \vspace{-0.8em}
  \includegraphics[width=\linewidth]{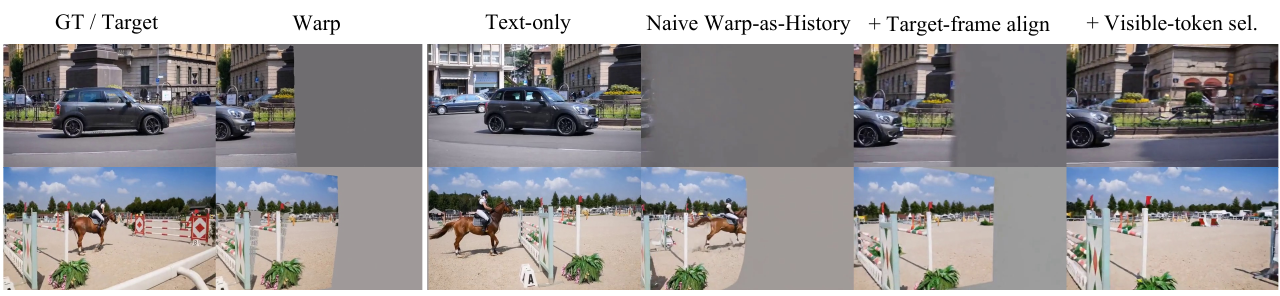}
  \vspace{-1.2em}
  \caption{Zero-shot interface ablation with the frozen model.}
  \label{fig:interface-analysis}
  \vspace{-0.8em}
\end{figure}

\paragraph{Interface ablation.}
Table~\ref{tab:activation-chain} compares NoAlign, NoVisDrop, and the full interface in both zero-shot and one-shot regimes.
Figure~\ref{fig:interface-analysis} visualizes the frozen-model chain: native warp history, target-frame positional alignment, and visible-token selection.

\begin{table}[H]
  \caption{Interface ablations on DAVIS and RE10K. VisLPIPS is visible-region LPIPS; Dyn. and Img. are VBench Dynamic Degree and Imaging Quality.}
  \label{tab:activation-chain}
  \centering
  \scriptsize
  \setlength{\tabcolsep}{2pt}
  \resizebox{\linewidth}{!}{%
  \begin{tabular}{llcccccccccccc}
    \toprule
    & & \multicolumn{6}{c}{DAVIS} & \multicolumn{6}{c}{RE10K} \\
    \cmidrule(lr){3-8}\cmidrule(lr){9-14}
    Regime & Setting & R-Err $\downarrow$ & T-Err $\downarrow$ & VisLPIPS $\downarrow$ & DOVER $\uparrow$ & Dyn. $\uparrow$ & Img. $\uparrow$ & R-Err $\downarrow$ & T-Err $\downarrow$ & VisLPIPS $\downarrow$ & DOVER $\uparrow$ & Dyn. $\uparrow$ & Img. $\uparrow$ \\
    \midrule
    Zero-shot & NoAlign & 7.33 & 0.1343 & 0.343 & \textbf{0.384} & \underline{0.740} & \textbf{58.7} & 4.26 & 0.1109 & 0.469 & 0.390 & \underline{0.980} & 59.0 \\
    & NoVisDrop & \underline{4.37} & \textbf{0.1151} & \underline{0.275} & 0.341 & \textbf{0.779} & 56.1 & \textbf{2.08} & \underline{0.0589} & \underline{0.249} & \underline{0.391} & \textbf{1.000} & \underline{62.4} \\
    & Full & \textbf{3.41} & \underline{0.1336} & \textbf{0.274} & \underline{0.364} & \textbf{0.779} & \underline{57.3} & \underline{2.20} & \textbf{0.0579} & \textbf{0.236} & \textbf{0.399} & \textbf{1.000} & \textbf{62.5} \\
    \midrule
    One-shot & ChFusion & 9.10 & 0.1902 & 0.289 & \textbf{0.478} & \underline{0.831} & \textbf{65.4} & 5.38 & 0.2178 & 0.257 & 0.422 & 0.790 & \underline{66.7} \\
    & SeqConcat & 5.86 & 0.1339 & 0.318 & 0.346 & 0.662 & 56.6 & 3.20 & 0.1451 & 0.331 & 0.241 & 0.710 & 41.3 \\
    & NoAlign & 7.06 & 0.1128 & 0.312 & \underline{0.471} & \textbf{0.896} & \underline{63.1} & 4.26 & 0.0806 & 0.392 & \underline{0.445} & \textbf{1.000} & 66.3 \\
    & NoVisDrop & \textbf{2.94} & \underline{0.1111} & \underline{0.231} & 0.452 & 0.753 & 62.5 & \underline{1.44} & \underline{0.0548} & \underline{0.157} & \textbf{0.449} & \underline{0.990} & \textbf{67.5} \\
    & Full & \underline{2.97} & \textbf{0.0942} & \textbf{0.224} & 0.448 & 0.714 & 62.2 & \textbf{1.28} & \textbf{0.0454} & \textbf{0.143} & 0.442 & 0.980 & 66.0 \\
    \bottomrule
  \end{tabular}
  }
\end{table}

\vspace{-1.0em}

\subsection{Small-Data Sensitivity}
\label{sec:oneshot-source-analysis}
\label{sec:few-shot-sensitivity}

We treat one-shot source selection and few-shot scaling as diagnostics rather than extra method components.
The fixed \emph{car-roundabout} source was chosen before the retrospective sweep; Appendix~\ref{sec:supp-oneshot-source} shows that it is high-performing but not the best source overall.
Appendix~\ref{app:additional-data-sensitivity} further shows that the clearest gain comes from zero-shot invocation to one-training-video finetuning, while adding more videos gives smaller, non-monotonic changes.
Thus the main evidence is not test-set-tuned source selection or many-video scaling, but that one separate video already activates the camera-follow prior across unseen videos.

\section{Conclusion}

We presented Warp-as-History, a one-training-video camera-control method that routes camera-warped pseudo-history through the native history pathway with target-frame positional alignment and visible-token selection.
The same interface reveals zero-shot camera-following behavior in a frozen model, and a lightweight LoRA update on one separate video stabilizes it without test-time optimization or per-video fitting.
Across WorldScore, RE10K, and DAVIS, the method is competitive with larger camera-control systems while preserving the pretrained backbone's generative behavior.

\bibliographystyle{plainnat}
\bibliography{references}

\begin{thebibliography}{27}
\providecommand{\natexlab}[1]{#1}
\providecommand{\url}[1]{\texttt{#1}}
\expandafter\ifx\csname urlstyle\endcsname\relax
  \providecommand{\doi}[1]{doi: #1}\else
  \providecommand{\doi}{doi: \begingroup \urlstyle{rm}\Url}\fi

\bibitem[Dai et~al.(2025)Dai, Jiang, Wang, Xu, and Qi]{dai2025fantasyworld}
Yixiang Dai, Fan Jiang, Chiyu Wang, Mu~Xu, and Yonggang Qi.
\newblock {FantasyWorld}: Geometry-consistent world modeling via unified video
  and 3d prediction.
\newblock \emph{arXiv preprint arXiv:2509.21657}, 2025.

\bibitem[Duan et~al.(2025)Duan, Yu, Chen, Fei-Fei, and Wu]{duan2025worldscore}
Haoyi Duan, Hong-Xing Yu, Sirui Chen, Li~Fei-Fei, and Jiajun Wu.
\newblock {WorldScore}: A unified evaluation benchmark for world generation.
\newblock In \emph{Proceedings of the IEEE/CVF International Conference on
  Computer Vision}, pages 27713--27724, 2025.

\bibitem[He et~al.(2024)He, Xu, Guo, Wetzstein, Dai, Li, and
  Yang]{he2024cameractrl}
Hao He, Yinghao Xu, Yuwei Guo, Gordon Wetzstein, Bo~Dai, Hongsheng Li, and
  Ceyuan Yang.
\newblock {CameraCtrl}: Enabling camera control for text-to-video generation.
\newblock \emph{arXiv preprint arXiv:2404.02101}, 2024.

\bibitem[Hou and Chen(2024)]{hou2024training}
Chen Hou and Zhibo Chen.
\newblock Training-free camera control for video generation.
\newblock \emph{arXiv preprint arXiv:2406.10126}, 2024.

\bibitem[Hu et~al.(2022)Hu, Shen, Wallis, Allen-Zhu, Li, Wang, Wang, Chen,
  et~al.]{hu2022lora}
Edward~J Hu, Yelong Shen, Phillip Wallis, Zeyuan Allen-Zhu, Yuanzhi Li, Shean
  Wang, Liang Wang, Weizhu Chen, et~al.
\newblock Lora: Low-rank adaptation of large language models.
\newblock \emph{Iclr}, 1\penalty0 (2):\penalty0 3, 2022.

\bibitem[Huang et~al.(2025{\natexlab{a}})Huang, Zheng, Wang, Liu, Wang, Wu,
  Jiang, Li, Lau, Zuo, et~al.]{huang2025voyager}
Tianyu Huang, Wangguandong Zheng, Tengfei Wang, Yuhao Liu, Zhenwei Wang, Junta
  Wu, Jie Jiang, Hui Li, Rynson Lau, Wangmeng Zuo, et~al.
\newblock Voyager: Long-range and world-consistent video diffusion for
  explorable 3d scene generation.
\newblock \emph{ACM Transactions on Graphics (TOG)}, 44\penalty0 (6):\penalty0
  1--15, 2025{\natexlab{a}}.

\bibitem[Huang et~al.(2025{\natexlab{b}})Huang, Li, He, Zhou, and
  Shechtman]{huang2025self}
Xun Huang, Zhengqi Li, Guande He, Mingyuan Zhou, and Eli Shechtman.
\newblock Self forcing: Bridging the train-test gap in autoregressive video
  diffusion.
\newblock \emph{arXiv preprint arXiv:2506.08009}, 2025{\natexlab{b}}.

\bibitem[Huang et~al.(2024)Huang, He, Yu, Zhang, Si, Jiang, Zhang, Wu, Jin,
  Chanpaisit, et~al.]{huang2024vbench}
Ziqi Huang, Yinan He, Jiashuo Yu, Fan Zhang, Chenyang Si, Yuming Jiang, Yuanhan
  Zhang, Tianxing Wu, Qingyang Jin, Nattapol Chanpaisit, et~al.
\newblock Vbench: Comprehensive benchmark suite for video generative models.
\newblock In \emph{Proceedings of the IEEE/CVF Conference on Computer Vision
  and Pattern Recognition}, pages 21807--21818, 2024.

\bibitem[Li et~al.(2025)Li, Yi, Liu, Gao, Ma, and Kanazawa]{li2025cameras}
Ruilong Li, Brent Yi, Junchen Liu, Hang Gao, Yi~Ma, and Angjoo Kanazawa.
\newblock Cameras as relative positional encoding.
\newblock \emph{arXiv preprint arXiv:2507.10496}, 2025.

\bibitem[Liu et~al.(2024)Liu, Shao, and Lu]{liu2024novel}
Kunhao Liu, Ling Shao, and Shijian Lu.
\newblock Novel view extrapolation with video diffusion priors.
\newblock \emph{arXiv preprint arXiv:2411.14208}, 2024.

\bibitem[Perazzi et~al.(2016)Perazzi, Pont-Tuset, McWilliams, Van~Gool, Gross,
  and Sorkine-Hornung]{perazzi2016benchmark}
Federico Perazzi, Jordi Pont-Tuset, Brian McWilliams, Luc Van~Gool, Markus
  Gross, and Alexander Sorkine-Hornung.
\newblock A benchmark dataset and evaluation methodology for video object
  segmentation.
\newblock In \emph{Proceedings of the IEEE Conference on Computer Vision and
  Pattern Recognition}, pages 724--732, 2016.

\bibitem[Ren et~al.(2025)Ren, Shen, Huang, Ling, Lu, Nimier-David, M{\"u}ller,
  Keller, Fidler, and Gao]{ren2025gen3c}
Xuanchi Ren, Tianchang Shen, Jiahui Huang, Huan Ling, Yifan Lu, Merlin
  Nimier-David, Thomas M{\"u}ller, Alexander Keller, Sanja Fidler, and Jun Gao.
\newblock Gen3c: 3d-informed world-consistent video generation with precise
  camera control.
\newblock In \emph{Proceedings of the IEEE/CVF Conference on Computer Vision
  and Pattern Recognition}, pages 6121--6132, 2025.

\bibitem[Song et~al.(2025{\natexlab{a}})Song, Yang, Zhao, Li, and
  Zhang]{song2025worldforge}
Chenxi Song, Yanming Yang, Tong Zhao, Ruibo Li, and Chi Zhang.
\newblock {WorldForge}: Unlocking emergent 3d/4d generation in video diffusion
  model via training-free guidance.
\newblock \emph{arXiv preprint arXiv:2509.15130}, 2025{\natexlab{a}}.

\bibitem[Song et~al.(2025{\natexlab{b}})Song, Chen, Simchowitz, Du, Tedrake,
  and Sitzmann]{song2025history}
Kiwhan Song, Boyuan Chen, Max Simchowitz, Yilun Du, Russ Tedrake, and Vincent
  Sitzmann.
\newblock History-guided video diffusion.
\newblock \emph{arXiv preprint arXiv:2502.06764}, 2025{\natexlab{b}}.

\bibitem[Su et~al.(2024)Su, Ahmed, Lu, Pan, Bo, and Liu]{su2024roformer}
Jianlin Su, Murtadha Ahmed, Yu~Lu, Shengfeng Pan, Wen Bo, and Yunfeng Liu.
\newblock Roformer: Enhanced transformer with rotary position embedding.
\newblock \emph{Neurocomputing}, 568:\penalty0 127063, 2024.

\bibitem[Sun et~al.(2025)Sun, Zhang, Wang, Wu, Wang, Wang, Wang, Zhang, Wang,
  and Guo]{sun2025worldplay}
Wenqiang Sun, Haiyu Zhang, Haoyuan Wang, Junta Wu, Zehan Wang, Zhenwei Wang,
  Yunhong Wang, Jun Zhang, Tengfei Wang, and Chunchao Guo.
\newblock Worldplay: Towards long-term geometric consistency for real-time
  interactive world modeling.
\newblock \emph{arXiv preprint arXiv:2512.14614}, 2025.

\bibitem[Wang et~al.(2025)Wang, Zhou, Zhu, Chang, Zhou, Li, Chen, Pang, Shen,
  and He]{wang2025pi}
Yifan Wang, Jianjun Zhou, Haoyi Zhu, Wenzheng Chang, Yang Zhou, Zizun Li, Junyi
  Chen, Jiangmiao Pang, Chunhua Shen, and Tong He.
\newblock $\pi^3$: Permutation-equivariant visual geometry learning.
\newblock \emph{arXiv preprint arXiv:2507.13347}, 2025.

\bibitem[Wu et~al.(2023)Wu, Zhang, Liao, Chen, Hou, Wang, Sun, Yan, and
  Lin]{wu2023exploring}
Haoning Wu, Erli Zhang, Liang Liao, Chaofeng Chen, Jingwen Hou, Annan Wang,
  Wenxiu Sun, Qiong Yan, and Weisi Lin.
\newblock Exploring video quality assessment on user generated contents from
  aesthetic and technical perspectives.
\newblock In \emph{Proceedings of the IEEE/CVF International Conference on
  Computer Vision}, pages 20144--20154, 2023.

\bibitem[Wu et~al.(2025)Wu, Yang, Po, Xu, Liu, Lin, and Wetzstein]{wu2025video}
Tong Wu, Shuai Yang, Ryan Po, Yinghao Xu, Ziwei Liu, Dahua Lin, and Gordon
  Wetzstein.
\newblock Video world models with long-term spatial memory.
\newblock \emph{arXiv preprint arXiv:2506.05284}, 2025.

\bibitem[Yang et~al.(2024)Yang, Teng, Zheng, Ding, Huang, Xu, Yang, Hong,
  Zhang, Feng, et~al.]{yang2024cogvideox}
Zhuoyi Yang, Jiayan Teng, Wendi Zheng, Ming Ding, Shiyu Huang, Jiazheng Xu,
  Yuanming Yang, Wenyi Hong, Xiaohan Zhang, Guanyu Feng, et~al.
\newblock {CogVideoX}: Text-to-video diffusion models with an expert
  transformer.
\newblock \emph{arXiv preprint arXiv:2408.06072}, 2024.

\bibitem[You et~al.(2024)You, Zhu, Liu, and Hou]{you2024nvs}
Meng You, Zhiyu Zhu, Hui Liu, and Junhui Hou.
\newblock {NVS-Solver}: Video diffusion model as zero-shot novel view
  synthesizer.
\newblock \emph{arXiv preprint arXiv:2405.15364}, 2024.

\bibitem[Yu et~al.(2025)Yu, Bai, Qin, Liu, Wang, Wan, Zhang, and
  Liu]{yu2025context}
Jiwen Yu, Jianhong Bai, Yiran Qin, Quande Liu, Xintao Wang, Pengfei Wan,
  Di~Zhang, and Xihui Liu.
\newblock Context as memory: Scene-consistent interactive long video generation
  with memory retrieval.
\newblock In \emph{Proceedings of the SIGGRAPH Asia 2025 Conference Papers},
  pages 1--11, 2025.

\bibitem[Yu et~al.(2024)Yu, Xing, Yuan, Hu, Li, Huang, Gao, Wong, Shan, and
  Tian]{yu2024viewcrafter}
Wangbo Yu, Jinbo Xing, Li~Yuan, Wenbo Hu, Xiaoyu Li, Zhipeng Huang, Xiangjun
  Gao, Tien-Tsin Wong, Ying Shan, and Yonghong Tian.
\newblock Viewcrafter: Taming video diffusion models for high-fidelity novel
  view synthesis.
\newblock \emph{arXiv preprint arXiv:2409.02048}, 2024.

\bibitem[Yuan et~al.(2026)Yuan, Yin, Li, Huang, Yang, and Yuan]{yuan2026helios}
Shenghai Yuan, Yuanyang Yin, Zongjian Li, Xinwei Huang, Xiao Yang, and Li~Yuan.
\newblock Helios: Real real-time long video generation model.
\newblock \emph{arXiv preprint arXiv:2603.04379}, 2026.

\bibitem[Zhang et~al.(2025)Zhang, Li, Wei, Cao, Gambardella, Phung, and
  Cai]{zhang2025unified}
Cheng Zhang, Boying Li, Meng Wei, Yan-Pei Cao, Camilo~Cruz Gambardella, Dinh
  Phung, and Jianfei Cai.
\newblock Unified camera positional encoding for controlled video generation.
\newblock \emph{arXiv preprint arXiv:2512.07237}, 2025.

\bibitem[Zhou et~al.(2018)Zhou, Tucker, Flynn, Fyffe, and
  Snavely]{zhou2018stereo}
Tinghui Zhou, Richard Tucker, John Flynn, Graham Fyffe, and Noah Snavely.
\newblock Stereo magnification: Learning view synthesis using multiplane
  images.
\newblock \emph{arXiv preprint arXiv:1805.09817}, 2018.

\bibitem[Zhou et~al.(2025)Zhou, An, and Luo]{zhou2025latent}
Zhenghong Zhou, Jie An, and Jiebo Luo.
\newblock {Latent-Reframe}: Enabling camera control for video diffusion models
  without training.
\newblock In \emph{Proceedings of the IEEE/CVF International Conference on
  Computer Vision}, pages 12779--12789, 2025.

\end{thebibliography}

\appendix

\section{HyWorldPlay Evaluation Details}
\label{app:hyworldplay-details}

For Table~\ref{tab:hyworldplay-vbench}, we randomly sample 50 images from WorldScore.
For each image, we sample three random camera directions, generate 30-second videos, and evaluate the results with VBench.
This protocol is separate from the native 33-frame WorldScore evaluation in Table~\ref{tab:worldscore-video-models}.

\section{Additional Qualitative Comparisons}
\label{app:additional-qualitative-comparisons}

\begin{figure}[H]
  \centering
  \includegraphics[width=\linewidth]{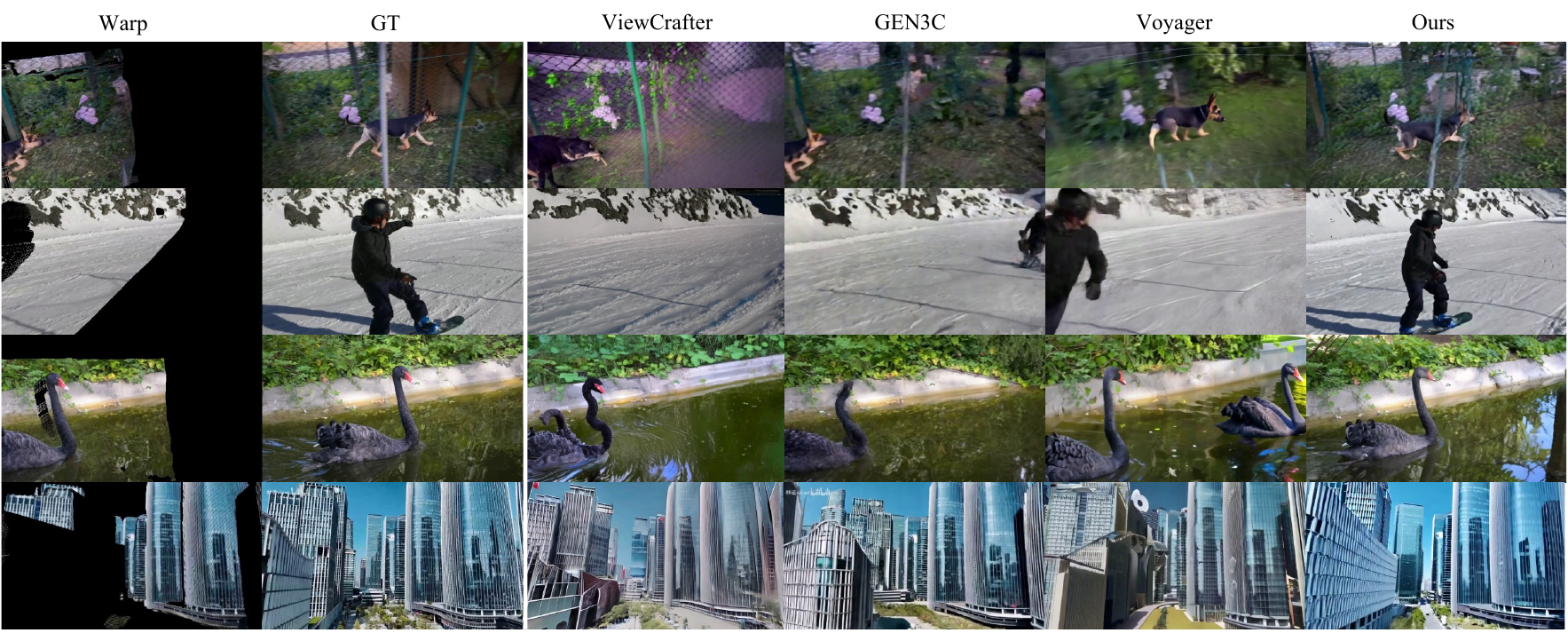}
  \vspace{-0.6em}
  \caption{Additional qualitative comparison with external camera-control methods on in-the-wild videos. Columns follow the same layout as Figure~\ref{fig:external-qualitative-comparison}.}
  \label{fig:external-qualitative-comparison-supp}
  \vspace{-0.8em}
\end{figure}

\section{Interface Ablation Settings and Full Tables}
\label{app:full-interface-ablation-tables}

The main text reports compact interface-ablation results in Table~\ref{tab:activation-chain} and external-baseline results in Tables~\ref{tab:external-geometry} and~\ref{tab:external-quality}.
Here we provide the exact evaluation settings, full interface-ablation tables, and auxiliary external-baseline metrics omitted from the compact main tables.

\paragraph{Training and sequence selection.}
Training videos are sampled from sequences disjoint from all evaluation videos.
When evaluation is reported on a subset for compute reasons, the subset is randomly selected once before evaluation and then fixed for all compared rows in that table.
Thus every method in a table is evaluated on the same videos and target camera trajectories.

\paragraph{Implementation details.}
For LoRA finetuning, we train LoRA parameters on the Helios-Mid checkpoint and mount the resulting update on the distilled checkpoint at inference time.
The distilled inference model uses six denoising steps grouped as $2+2+2$ across three resolution stages, with the first step of each later stage upsampling noise tokens from the previous stage.
Aligned warp history and LoRA are applied only in the first stage; later higher-resolution stages are left unchanged.
Within the first stage, LoRA adapters are mounted on the self-attention query, key, value, and output projections with rank 32 and scaling factor $\alpha=32$.
For camera-warped pseudo-history, we do not apply the temporal compression used by ordinary history; the warped video is kept at the original VAE latent resolution before being patchified and inserted into the history stream.
The current main recipe uses a generic camera-control trigger, trains for 1000 iterations on one source video, and uses no test-time optimization.

\paragraph{Shared row definitions.}
For the interface-ablation tables, all rows use Helios-Distilled as the base inference model.
The text-only baseline receives no camera-control condition at inference time, but camera metrics are still computed against the same target camera reference used by the camera-conditioned rows.
\texttt{Full} denotes warp-as-history with RoPE alignment and visible-token dropping; \texttt{NoAlign} removes RoPE alignment; \texttt{NoVisDrop} keeps invisible warp tokens; \texttt{SeqConcat} and \texttt{ChFusion} are non-history conditioning baselines.
\texttt{ChFusion} is a Gen3C-like channel-fusion baseline: the warped target-view latents are concatenated with the noisy target latents along the channel dimension and fused before denoising, rather than being routed through the history stream.
\texttt{SeqConcat} is a sequence-concatenation baseline: warp tokens are appended to the denoising-token sequence as ordinary condition tokens, following the same sequence-level path as the noisy target tokens instead of the pretrained history pathway.
Zero-shot rows use no LoRA, while one-shot rows use the same stage0-only distilled inference protocol with the corresponding one-shot LoRA update.

\paragraph{WorldScore setting.}
The WorldScore~\citep{duan2025worldscore} report uses the \texttt{static\_cc\_dev32} subset under native 33-frame evaluation.
It contains 32 deterministic samples selected from metadata before evaluation: one sample for each combination of 2 visual styles, 2 scene types, and 8 single-camera motions.
All listed rows are evaluated on the same 32 samples.

\paragraph{DAVIS setting.}
The DAVIS~\citep{perazzi2016benchmark} report uses a 77-video, common-33-frame first-chunk protocol.
It contains 77 DAVIS videos with at least 33 common frames; videos without enough frames are excluded.
For each video, the evaluated window starts at frame 0 and uses the first 33-frame chunk.
Camera-control rows use Pi3X-estimated pseudo-ground-truth camera trajectories from the original DAVIS videos as the target camera condition; if a model requires intrinsics, it uses the paired Pi3X intrinsics for the same 33 frames.
Evaluation follows the same DAVIS camera-control protocol for all compared rows.

\paragraph{RE10K setting.}
The RE10K~\citep{zhou2018stereo} report uses a fixed 100-sequence subset of the test split for the DAVIS-aligned ablation.
It follows the same row definitions as the DAVIS table and reports the available camera, DOVER, and VBench axes.
Metrics unavailable in the source report are omitted rather than filled with proxy values.

\paragraph{DAVIS external baseline setting.}
The external-baseline report compares Ours (one-shot) against Gen3C~\citep{ren2025gen3c}, Voyager~\citep{huang2025voyager} with the corrected crop, and ViewCrafter~\citep{yu2024viewcrafter} on DAVIS.
The Ours row uses Warp-as-History after one-training-video finetuning, while the external rows use the matched baseline outputs.
All rows use the common 33-frame DAVIS evaluation and the same camera, DOVER, and VBench report card.

\paragraph{RE10K external baseline setting.}
The RE10K external-baseline report compares Ours (one-shot) against Gen3C~\citep{ren2025gen3c}, Voyager~\citep{huang2025voyager}, and ViewCrafter~\citep{yu2024viewcrafter} on the same 99 RE10K sequences.
One sequence is excluded because the corresponding ViewCrafter output is unavailable.
Camera metrics use 33 frames with Pi3X frame stride 4, and all rows share the same camera, DOVER, and VBench report card.

\paragraph{Additional-data sensitivity.}
\label{app:additional-data-sensitivity}
The main paper centers on one-training-video finetuning.
We include the multi-video runs only as a sensitivity check, because increasing the training set in the current range, up to 12 videos, does not produce a clear monotonic improvement over the one-training-video setting.
Table~\ref{tab:fewshot-sensitivity} reports DAVIS+RE10K mean metrics, while Tables~\ref{tab:supp-fewshot-davis} and~\ref{tab:supp-fewshot-re10k} provide the full per-dataset report cards.

\begin{table}[H]
  \caption{Few-shot sensitivity (DAVIS+RE10K mean). The 0 row is zero-shot without LoRA.}
  \label{tab:fewshot-sensitivity}
  \centering
  \footnotesize
  \setlength{\tabcolsep}{4pt}
  \begin{tabular}{@{}lcccccc@{}}
    \toprule
    Tuning videos & PSNR $\uparrow$ & LPIPS $\downarrow$ & R-Err $\downarrow$ & T-Err $\downarrow$ & DOVER $\uparrow$ & Img. $\uparrow$ \\
    \midrule
    0 & 13.38 & 0.4178 & 2.81 & 0.0958 & 0.381 & 59.93 \\
    \midrule
    1 & \textbf{16.02} & \underline{0.3136} & 2.25 & \underline{0.0766} & 0.447 & 64.47 \\
    3 & \underline{15.89} & \textbf{0.3115} & \underline{1.84} & 0.0779 & 0.461 & 65.09 \\
    5 & 15.04 & 0.3564 & 1.90 & 0.0851 & 0.471 & 65.71 \\
    7 & 15.63 & 0.3163 & \textbf{1.72} & \textbf{0.0751} & \underline{0.475} & \underline{65.76} \\
    10 & 15.49 & 0.3242 & 1.99 & 0.0772 & 0.473 & 65.30 \\
    12 & 15.08 & 0.3483 & 2.12 & 0.0807 & \textbf{0.478} & \textbf{65.92} \\
    \bottomrule
  \end{tabular}
\end{table}

\begin{table}[H]
  \caption{DAVIS few-shot sensitivity full metrics. Rows differ in the number of source videos used for LoRA finetuning under the same small-data recipe.}
  \label{tab:supp-fewshot-davis}
  \centering
  \scriptsize
  \setlength{\tabcolsep}{1.8pt}
  \resizebox{\linewidth}{!}{%
  \begin{tabular}{lccccccccccccccccc}
    \toprule
    Videos & PSNR $\uparrow$ & SSIM $\uparrow$ & LPIPS $\downarrow$ & VisLPIPS $\downarrow$ & R-Err $\downarrow$ & T-Err $\downarrow$ & FID $\downarrow$ & FVD $\downarrow$ & DOVER $\uparrow$ & Tech. $\uparrow$ & Aesth. $\uparrow$ & Flicker $\uparrow$ & Motion $\uparrow$ & Subject $\uparrow$ & Backgr. $\uparrow$ & Dyn. $\uparrow$ & Img. $\uparrow$ \\
    \midrule
    1 & 15.14 & 0.3949 & 0.3820 & 0.2252 & 3.06 & 0.1076 & 67.37 & 58.52 & 0.453 & 0.0760 & 0.9509 & 0.9397 & 0.9765 & 0.9403 & 0.9395 & 0.714 & 62.26 \\
    3 & 15.02 & 0.3921 & 0.3737 & 0.2219 & 2.54 & 0.1051 & 64.00 & 49.48 & 0.472 & 0.0797 & 0.9617 & 0.9360 & 0.9754 & 0.9279 & 0.9340 & 0.831 & 61.93 \\
    5 & 14.59 & 0.3780 & 0.3911 & 0.2340 & 2.18 & 0.1013 & 64.43 & 48.70 & 0.472 & 0.0799 & 0.9620 & 0.9332 & 0.9734 & 0.9198 & 0.9303 & 0.883 & 61.93 \\
    7 & 14.66 & 0.3802 & 0.3815 & 0.2262 & 2.27 & 0.1055 & 63.15 & 49.32 & 0.478 & 0.0813 & 0.9630 & 0.9354 & 0.9759 & 0.9228 & 0.9310 & 0.818 & 61.76 \\
    10 & 14.55 & 0.3763 & 0.3907 & 0.2367 & 2.53 & 0.1052 & 66.07 & 51.02 & 0.479 & 0.0802 & 0.9635 & 0.9320 & 0.9728 & 0.9162 & 0.9313 & 0.883 & 62.09 \\
    12 & 14.41 & 0.3723 & 0.3991 & 0.2417 & 2.68 & 0.1083 & 65.13 & 54.68 & 0.481 & 0.0809 & 0.9634 & 0.9320 & 0.9740 & 0.9102 & 0.9291 & 0.961 & 61.94 \\
    \bottomrule
  \end{tabular}
  }
\end{table}

\begin{table}[H]
  \caption{RE10K few-shot sensitivity full metrics. The same LoRA updates as Table~\ref{tab:supp-fewshot-davis} are evaluated on RE10K without per-test-video adaptation.}
  \label{tab:supp-fewshot-re10k}
  \centering
  \scriptsize
  \setlength{\tabcolsep}{1.8pt}
  \resizebox{\linewidth}{!}{%
  \begin{tabular}{lccccccccccccccccc}
    \toprule
    Videos & PSNR $\uparrow$ & SSIM $\uparrow$ & LPIPS $\downarrow$ & VisLPIPS $\downarrow$ & R-Err $\downarrow$ & T-Err $\downarrow$ & FID $\downarrow$ & FVD $\downarrow$ & DOVER $\uparrow$ & Tech. $\uparrow$ & Aesth. $\uparrow$ & Flicker $\uparrow$ & Motion $\uparrow$ & Subject $\uparrow$ & Backgr. $\uparrow$ & Dyn. $\uparrow$ & Img. $\uparrow$ \\
    \midrule
    1 & 16.91 & 0.6108 & 0.2453 & 0.1514 & 1.45 & 0.0456 & 21.96 & 17.88 & 0.442 & 0.0696 & 0.9639 & 0.9540 & 0.9910 & 0.9571 & 0.9587 & 0.980 & 66.68 \\
    3 & 16.77 & 0.6098 & 0.2492 & 0.1558 & 1.15 & 0.0506 & 34.54 & 30.37 & 0.451 & 0.0714 & 0.9663 & 0.9510 & 0.9900 & 0.9441 & 0.9491 & 0.990 & 68.24 \\
    5 & 15.50 & 0.5395 & 0.3216 & 0.2068 & 1.61 & 0.0688 & 33.21 & 30.32 & 0.471 & 0.0738 & 0.9701 & 0.9474 & 0.9889 & 0.9409 & 0.9485 & 0.990 & 69.49 \\
    7 & 16.61 & 0.5972 & 0.2511 & 0.1553 & 1.17 & 0.0448 & 30.40 & 29.64 & 0.471 & 0.0741 & 0.9703 & 0.9506 & 0.9903 & 0.9484 & 0.9509 & 0.990 & 69.76 \\
    10 & 16.43 & 0.6004 & 0.2577 & 0.1641 & 1.45 & 0.0493 & 24.81 & 22.43 & 0.466 & 0.0730 & 0.9700 & 0.9498 & 0.9904 & 0.9477 & 0.9530 & 1.000 & 68.51 \\
    12 & 15.75 & 0.5705 & 0.2976 & 0.1952 & 1.56 & 0.0531 & 43.98 & 38.81 & 0.474 & 0.0729 & 0.9731 & 0.9472 & 0.9894 & 0.9363 & 0.9460 & 0.970 & 69.89 \\
    \bottomrule
  \end{tabular}
  }
\end{table}

\section{Supplementary One-Shot Source Diagnostics}
\label{sec:supp-oneshot-source}

Table~\ref{tab:oneshot-source-sensitivity} reports a compact subset of the one-shot source sweep.
The full sweep records PSNR, SSIM, LPIPS, visible-region LPIPS, R-Err, T-Err, FID/FVD, DOVER, and VBench axes for each source clip.
Tables~\ref{tab:supp-oneshot-source-davis} and~\ref{tab:supp-oneshot-source-re10k} report the main reconstruction, camera-following, distribution, and quality axes for the full one-shot source sweep.
For each source clip, we also profile the mean/max invisible ratio, mean/max camera rotation, translation-direction angle, foreground centroid motion and area, and source/warp visual quality.
The useful criterion is balanced, stable camera-induced parallax rather than raw motion magnitude or the largest invisible region.
Effective sources provide clear camera-induced parallax, moderate and stable disocclusion, limited foreground self-motion, and clean source/warp visual quality.
The omitted source rows follow the same pattern as the compact table: weak camera-signal clips such as \emph{breakdance} and unstable foreground/warp cases underperform even though the LoRA recipe is unchanged.
For example, \emph{breakdance} and \emph{drift-chicane} have very low invisible ratios but almost no camera rotation, so they provide a weak camera-control signal; \emph{motocross-bumps} and \emph{bmx-bumps} contain larger motion and invisible regions, but their unstable motion and disocclusion are not consistently beneficial.

\begin{table}[H]
  \caption{One-shot source sensitivity. The LoRA recipe is fixed and only the source video changes. Mean rank aggregates the full DAVIS/RE10K source sweep, with lower values better.}
  \label{tab:oneshot-source-sensitivity}
  \centering
  \scriptsize
  \setlength{\tabcolsep}{4pt}
  \resizebox{\linewidth}{!}{%
  \begin{tabular}{@{}l|ccc|ccc@{}}
    \toprule
    Source & \multicolumn{3}{c|}{Source diagnostics} & \multicolumn{3}{c}{Downstream results} \\
    \cmidrule(lr){2-4}\cmidrule(l){5-7}
    & Inv. mean & Rot. mean & Fg motion & DAVIS LPIPS $\downarrow$ & RE10K LPIPS $\downarrow$ & Mean rank $\downarrow$ \\
    \midrule
    \emph{train} & 0.105 & 2.82$^\circ$ & 0.0046 & \textbf{0.372} & \textbf{0.228} & \textbf{2.13} \\
    \emph{horsejump-high} & 0.258 & 12.14$^\circ$ & 0.0098 & \underline{0.375} & 0.236 & \underline{3.06} \\
    \emph{car-roundabout} & 0.322 & 11.15$^\circ$ & 0.0024 & 0.377 & \underline{0.235} & 3.94 \\
    \emph{drift-chicane} & 0.014 & 0.03$^\circ$ & 0.0103 & 0.407 & 0.307 & 8.38 \\
    \emph{classic-car} & 0.252 & 29.18$^\circ$ & 0.0072 & 0.409 & 0.316 & 8.75 \\
    \emph{surf} & 0.228 & 1.45$^\circ$ & 0.0174 & 0.452 & 0.479 & 10.88 \\
    \bottomrule
  \end{tabular}
  }
\end{table}

\begin{table}[H]
  \caption{DAVIS one-shot source sweep. Rows differ only in the source video used for one-shot finetuning. VisLPIPS is visible-region LPIPS; Dyn. and Img. are VBench Dynamic Degree and Imaging Quality.}
  \label{tab:supp-oneshot-source-davis}
  \centering
  \scriptsize
  \setlength{\tabcolsep}{2.2pt}
  \resizebox{\linewidth}{!}{%
  \begin{tabular}{lccccccccccc}
    \toprule
    Source & PSNR $\uparrow$ & SSIM $\uparrow$ & LPIPS $\downarrow$ & VisLPIPS $\downarrow$ & R-Err $\downarrow$ & T-Err $\downarrow$ & FID $\downarrow$ & FVD $\downarrow$ & DOVER $\uparrow$ & Dyn. $\uparrow$ & Img. $\uparrow$ \\
    \midrule
    \emph{bmx-bumps} & 14.40 & 0.3668 & 0.4003 & 0.2378 & 2.89 & 0.0976 & 67.97 & 53.48 & 0.431 & 0.857 & 60.26 \\
    \emph{breakdance} & 14.62 & 0.3839 & 0.4145 & 0.2501 & 3.54 & 0.1110 & 71.26 & 57.76 & 0.433 & 0.909 & 59.90 \\
    \emph{car-roundabout} & 15.21 & 0.3980 & 0.3772 & 0.2224 & 3.03 & 0.1033 & 68.69 & 56.02 & 0.453 & 0.727 & 62.41 \\
    \emph{classic-car} & 14.64 & 0.3707 & 0.4085 & 0.2481 & 2.76 & 0.0905 & 73.64 & 64.07 & 0.415 & 0.779 & 59.86 \\
    \emph{dance-twirl} & 15.11 & 0.3970 & 0.3786 & 0.2264 & 2.87 & 0.1042 & 64.66 & 49.25 & 0.456 & 0.831 & 60.65 \\
    \emph{drift-chicane} & 15.04 & 0.3942 & 0.4069 & 0.2461 & 3.51 & 0.1038 & 68.13 & 56.87 & 0.427 & 0.740 & 60.16 \\
    \emph{horsejump-high} & 15.38 & 0.4073 & 0.3746 & 0.2225 & 2.76 & 0.0933 & 65.42 & 53.18 & 0.440 & 0.766 & 60.95 \\
    \emph{kite-surf} & 15.17 & 0.3959 & 0.3992 & 0.2389 & 3.06 & 0.1032 & 70.00 & 60.24 & 0.418 & 0.740 & 60.00 \\
    \emph{motocross-bumps} & 15.10 & 0.3936 & 0.3884 & 0.2311 & 2.60 & 0.1054 & 67.37 & 60.31 & 0.443 & 0.740 & 60.60 \\
    \emph{parkour} & 15.04 & 0.3870 & 0.3864 & 0.2333 & 2.91 & 0.1045 & 64.98 & 55.89 & 0.446 & 0.766 & 62.33 \\
    \emph{surf} & 14.08 & 0.3516 & 0.4515 & 0.2842 & 3.97 & 0.0998 & 71.38 & 61.29 & 0.413 & 0.805 & 59.28 \\
    \emph{train} & 15.31 & 0.4034 & 0.3717 & 0.2205 & 2.72 & 0.0968 & 68.59 & 55.81 & 0.447 & 0.753 & 61.39 \\
    \bottomrule
  \end{tabular}
  }
\end{table}

\begin{table}[H]
  \caption{RE10K one-shot source sweep. Rows use the same source-video LoRA updates as Table~\ref{tab:supp-oneshot-source-davis}, evaluated on RE10K without per-test-video adaptation.}
  \label{tab:supp-oneshot-source-re10k}
  \centering
  \scriptsize
  \setlength{\tabcolsep}{2.2pt}
  \resizebox{\linewidth}{!}{%
  \begin{tabular}{lccccccccccc}
    \toprule
    Source & PSNR $\uparrow$ & SSIM $\uparrow$ & LPIPS $\downarrow$ & VisLPIPS $\downarrow$ & R-Err $\downarrow$ & T-Err $\downarrow$ & FID $\downarrow$ & FVD $\downarrow$ & DOVER $\uparrow$ & Dyn. $\uparrow$ & Img. $\uparrow$ \\
    \midrule
    \emph{bmx-bumps} & 15.81 & 0.5658 & 0.2872 & 0.1828 & 1.43 & 0.0541 & 22.55 & 20.94 & 0.451 & 1.000 & 67.92 \\
    \emph{breakdance} & 15.78 & 0.5752 & 0.3024 & 0.1992 & 1.72 & 0.0610 & 50.22 & 46.21 & 0.407 & 1.000 & 63.93 \\
    \emph{car-roundabout} & 17.12 & 0.6187 & 0.2353 & 0.1426 & 1.32 & 0.0456 & 21.60 & 18.16 & 0.442 & 0.980 & 66.37 \\
    \emph{classic-car} & 15.48 & 0.5621 & 0.3162 & 0.2088 & 1.38 & 0.0613 & 23.09 & 19.71 & 0.433 & 1.000 & 65.73 \\
    \emph{dance-twirl} & 16.78 & 0.6074 & 0.2514 & 0.1582 & 1.31 & 0.0478 & 36.59 & 30.23 & 0.431 & 0.990 & 66.23 \\
    \emph{drift-chicane} & 15.86 & 0.5755 & 0.3069 & 0.2015 & 1.90 & 0.0554 & 24.47 & 21.29 & 0.417 & 0.990 & 63.62 \\
    \emph{horsejump-high} & 17.13 & 0.6274 & 0.2356 & 0.1466 & 1.29 & 0.0440 & 28.03 & 25.72 & 0.418 & 0.980 & 64.65 \\
    \emph{kite-surf} & 16.23 & 0.5883 & 0.2848 & 0.1827 & 1.65 & 0.0535 & 22.30 & 18.52 & 0.423 & 0.990 & 64.31 \\
    \emph{motocross-bumps} & 17.04 & 0.6220 & 0.2425 & 0.1485 & 1.28 & 0.0510 & 29.80 & 24.02 & 0.425 & 0.990 & 64.04 \\
    \emph{parkour} & 16.33 & 0.5858 & 0.2760 & 0.1790 & 1.56 & 0.0505 & 21.23 & 18.97 & 0.434 & 0.990 & 65.94 \\
    \emph{surf} & 13.31 & 0.4960 & 0.4788 & 0.3340 & 4.26 & 0.0693 & 23.50 & 23.34 & 0.416 & 1.000 & 61.51 \\
    \emph{train} & 17.22 & 0.6271 & 0.2282 & 0.1396 & 1.27 & 0.0481 & 21.21 & 17.45 & 0.442 & 0.980 & 66.49 \\
    \bottomrule
  \end{tabular}
  }
\end{table}

\begin{table}[H]
\caption{Additional DAVIS interface-ablation metrics from the same source as Table~\ref{tab:activation-chain}.}
  \label{tab:supp-davis-full}
  \centering
  \scriptsize
  \setlength{\tabcolsep}{2pt}
  \resizebox{\linewidth}{!}{%
  \begin{tabular}{llccccccccccccccc}
    \toprule
    Regime & Setting & PSNR $\uparrow$ & SSIM $\uparrow$ & LPIPS $\downarrow$ & Vis. LPIPS $\downarrow$ & R-Err $\downarrow$ & T-Err $\downarrow$ & FID $\downarrow$ & FVD $\downarrow$ & DOVER $\uparrow$ & Flicker $\uparrow$ & Motion $\uparrow$ & Subject $\uparrow$ & Backgr. $\uparrow$ & Dynamic $\uparrow$ & Imaging $\uparrow$ \\
    \midrule
    Text-only & Base & 14.38 & 0.3627 & 0.4199 & 0.2626 & 7.96 & 0.1968 & 71.55 & 80.87 & 0.463 & 0.989 & 0.994 & 0.979 & 0.971 & 0.091 & 65.21 \\
    \midrule
    Zero-shot & NoAlign & 12.03 & 0.2752 & 0.5439 & 0.3430 & 7.33 & 0.1343 & 94.86 & 83.98 & 0.384 & 0.948 & 0.973 & 0.912 & 0.929 & 0.740 & 58.75 \\
    & NoVisDrop & 12.31 & 0.3162 & 0.5013 & 0.2749 & 4.37 & 0.1151 & 89.69 & 72.28 & 0.341 & 0.957 & 0.978 & 0.901 & 0.920 & 0.779 & 56.05 \\
    & Full & 13.28 & 0.3441 & 0.4629 & 0.2735 & 3.41 & 0.1336 & 83.37 & 72.16 & 0.364 & 0.941 & 0.969 & 0.923 & 0.931 & 0.779 & 57.33 \\
    \midrule
    One-shot & ChFusion & 13.59 & 0.3150 & 0.4600 & 0.2891 & 9.10 & 0.1902 & 72.54 & 67.15 & 0.478 & 0.935 & 0.972 & 0.942 & 0.938 & 0.831 & 65.44 \\
    & SeqConcat & 12.42 & 0.3177 & 0.5019 & 0.3178 & 5.86 & 0.1339 & 100.52 & 90.31 & 0.346 & 0.942 & 0.969 & 0.919 & 0.937 & 0.662 & 56.59 \\
    & NoAlign & 13.47 & 0.3190 & 0.4826 & 0.3115 & 7.06 & 0.1128 & 69.62 & 64.32 & 0.471 & 0.936 & 0.975 & 0.927 & 0.932 & 0.896 & 63.08 \\
    & NoVisDrop & 15.02 & 0.3924 & 0.3863 & 0.2312 & 2.94 & 0.1111 & 68.58 & 56.87 & 0.452 & 0.940 & 0.976 & 0.937 & 0.935 & 0.753 & 62.54 \\
    & Full & 15.21 & 0.3976 & 0.3794 & 0.2236 & 2.97 & 0.0942 & 68.18 & 57.95 & 0.448 & 0.941 & 0.977 & 0.941 & 0.940 & 0.714 & 62.20 \\
    \bottomrule
  \end{tabular}
  }
\end{table}

\begin{table}[H]
\caption{Additional RE10K interface-ablation metrics from the same source as Table~\ref{tab:activation-chain}.}
  \label{tab:supp-re10k-full}
  \centering
  \scriptsize
  \setlength{\tabcolsep}{2pt}
  \resizebox{\linewidth}{!}{%
  \begin{tabular}{llccccccccccccccc}
    \toprule
    Regime & Setting & PSNR $\uparrow$ & SSIM $\uparrow$ & LPIPS $\downarrow$ & Vis. LPIPS $\downarrow$ & R-Err $\downarrow$ & T-Err $\downarrow$ & FID $\downarrow$ & FVD $\downarrow$ & DOVER $\uparrow$ & Flicker $\uparrow$ & Motion $\uparrow$ & Subject $\uparrow$ & Backgr. $\uparrow$ & Dynamic $\uparrow$ & Imaging $\uparrow$ \\
    \midrule
    Zero-shot & NoAlign & 9.46 & 0.3258 & 0.6834 & 0.4691 & 4.26 & 0.1109 & 85.95 & 68.27 & 0.390 & 0.964 & 0.991 & 0.933 & 0.952 & 0.980 & 58.96 \\
    & NoVisDrop & 13.09 & 0.5048 & 0.3906 & 0.2493 & 2.08 & 0.0589 & 36.81 & 29.79 & 0.391 & 0.960 & 0.988 & 0.953 & 0.955 & 1.000 & 62.40 \\
    & Full & 13.47 & 0.5220 & 0.3726 & 0.2358 & 2.20 & 0.0579 & 35.70 & 28.30 & 0.399 & 0.957 & 0.987 & 0.954 & 0.953 & 1.000 & 62.53 \\
    \midrule
    One-shot & ChFusion & 14.52 & 0.5247 & 0.3822 & 0.2566 & 5.38 & 0.2178 & 29.50 & 30.78 & 0.422 & 0.961 & 0.987 & 0.968 & 0.961 & 0.790 & 66.71 \\
    & SeqConcat & 12.24 & 0.5497 & 0.5289 & 0.3307 & 3.20 & 0.1451 & 94.80 & 82.01 & 0.241 & 0.973 & 0.991 & 0.911 & 0.944 & 0.710 & 41.25 \\
    & NoAlign & 12.37 & 0.4365 & 0.5585 & 0.3924 & 4.26 & 0.0806 & 33.89 & 35.58 & 0.445 & 0.950 & 0.989 & 0.936 & 0.948 & 1.000 & 66.28 \\
    & NoVisDrop & 16.78 & 0.6082 & 0.2523 & 0.1573 & 1.44 & 0.0548 & 21.61 & 18.33 & 0.449 & 0.953 & 0.990 & 0.956 & 0.957 & 0.990 & 67.50 \\
    & Full & 17.15 & 0.6214 & 0.2343 & 0.1426 & 1.28 & 0.0454 & 22.11 & 17.92 & 0.442 & 0.955 & 0.991 & 0.956 & 0.958 & 0.980 & 65.97 \\
    \bottomrule
  \end{tabular}
  }
\end{table}

\begin{table}[H]
\caption{Additional external-baseline metrics omitted from Tables~\ref{tab:external-geometry} and~\ref{tab:external-quality}.}
  \label{tab:supp-external-omitted-metrics}
  \centering
  \setlength{\tabcolsep}{4pt}
  \begin{tabular}{llccc}
    \toprule
    Dataset & Method & Visible \% & DOVER tech. $\uparrow$ & DOVER aesth. $\uparrow$ \\
    \midrule
    DAVIS & Ours & 66.0 & 0.0757 & 0.9479 \\
    & Gen3C & 66.0 & 0.0513 & 0.8954 \\
    & Voyager & 66.0 & 0.0649 & 0.9159 \\
    & ViewCrafter & 66.0 & 0.0799 & 0.9571 \\
    \midrule
    RE10K & Ours & 67.0 & 0.0697 & 0.9638 \\
    & Gen3C & 67.0 & 0.0409 & 0.9282 \\
    & Voyager & 67.0 & 0.0626 & 0.9446 \\
    & ViewCrafter & 67.0 & 0.0607 & 0.9658 \\
    \bottomrule
  \end{tabular}
\end{table}

\section{Runtime Analysis}
\label{app:runtime-analysis}

Warp-as-History adds camera-warped pseudo-history to the native history stream, so it inevitably increases the number of tokens processed by the transformer.
We measure runtime for generating one 33-frame chunk on a single NVIDIA A800 GPU.
Table~\ref{tab:supp-runtime-analysis} profiles this cost relative to the original image-to-video sampler under two visibility regimes.
The visible-token percentage denotes the fraction of warped-history tokens retained after visibility filtering; invisible-token dropping removes tokens without valid source observations before they enter the transformer.

The main overhead comes from the transformer/sampling stage rather than from geometry or warp preparation.
With 86\% visible tokens, transformer/sampling increases by 7.59s and accounts for almost all of the 7.81s end-to-end increase.
When only 47\% of the warped tokens are visible, the transformer/sampling overhead drops to 3.38s, and the end-to-end overhead drops to 4.62s.
By contrast, camera rendering, warp VAE encoding, and warp/mask preparation together contribute only about 1--2 seconds.
This confirms that Pi3X-based geometry estimation and warp preparation are not the bottleneck in this setting; the bottleneck is the longer transformer sequence induced by using warps as history.
The same trend also supports invisible-token dropping: discarding unobserved warp tokens not only avoids conditioning on invalid evidence, but also reduces the sequence length that dominates runtime.

\begin{table}[H]
  \caption{Runtime profile for generating one 33-frame chunk with Warp-as-History under different visible-token ratios on a single NVIDIA A800 GPU. Times are reported in seconds; \texttt{original} is the native image-to-video sampler and \texttt{ours} adds camera-warped pseudo-history.}
  \label{tab:supp-runtime-analysis}
  \centering
  \small
  \setlength{\tabcolsep}{5pt}
  \resizebox{\linewidth}{!}{%
  \begin{tabular}{llccc}
    \toprule
    Setting & Part & Original & Ours & Increase \\
    \midrule
    Warp with 86\% visible tokens & Camera render & 0.00s & 1.02s & +1.02s \\
    Warp with 86\% visible tokens & Warp VAE encode & 0.00s & 0.47s & +0.47s \\
    Warp with 86\% visible tokens & Transformer / sampling & 14.34s & 21.94s & +7.59s \\
    Warp with 86\% visible tokens & End-to-end & 15.83s & 23.63s & +7.81s \\
    \midrule
    Warp with 47\% visible tokens & Camera render, assumed & 0.00s & 1.02s & +1.02s \\
    Warp with 47\% visible tokens & Warp VAE encode & 0.00s & 0.47s & +0.47s \\
    Warp with 47\% visible tokens & Warp/mask prepare & 0.00s & 0.19s & +0.19s \\
    Warp with 47\% visible tokens & Transformer / sampling & 14.37s & 17.75s & +3.38s \\
    Warp with 47\% visible tokens & End-to-end & 15.78s & 20.40s & +4.62s \\
    \bottomrule
  \end{tabular}
  }
\end{table}

\section{Limitations}

Warp-as-History depends on the quality and cost of the warp construction step.
In our implementation, the warp is produced online by reconstructing the observed scene with an external reconstruction model and projecting the reconstruction to the target future cameras.
This avoids training a camera-specific control branch, but it adds preprocessing cost and inherits the reconstruction model's failure modes, including errors in geometry, visibility, and disoccluded regions.
The history interface itself is also not free: even though we insert warped history only in the first Helios stage, where the spatial resolution is low and inference uses only two denoising steps, the additional history tokens still increase runtime relative to the native image-to-video sampler.
Finally, the method is an invocation interface rather than a new video generator.
Its generalization is therefore bounded by the pretrained backbone's existing ability to interpret visual history, preserve dynamics, and complete unobserved content; when the base model lacks these capabilities, lightweight LoRA can stabilize the behavior but cannot fully remove the limitation.

\section{Broader Impacts}

This work studies camera control for pretrained video generation models.
It may be useful for creative editing, virtual cinematography, simulation, and controllable scene visualization.
Like other video generation and editing methods, it could also be misused to create misleading videos or to modify private or sensitive footage without consent.
Any release or deployment should therefore follow the safety policies of the underlying video model, clearly label generated or edited content, and respect dataset licenses and subject consent.
This paper does not introduce a new dataset or a deployed user-facing system.

\end{document}